\documentclass[sigconf]{acmart}
\usepackage{soul}
\usepackage{breqn}
\usepackage{multirow}
\usepackage{hyperref}
\usepackage{balance}
\AtBeginDocument{%
  \providecommand\BibTeX{{%
    \normalfont B\kern-0.5em{\scshape i\kern-0.25em b}\kern-0.8em\TeX}}}

\setcopyright{acmcopyright}
\copyrightyear{2022} 
\acmYear{2022} 
\setcopyright{acmcopyright}\acmConference[CIKM '22]{Proceedings of the 31st ACM International Conference on Information and Knowledge Management}{October 17--21, 2022}{Atlanta, GA, USA}
\acmBooktitle{Proceedings of the 31st ACM International Conference on Information and Knowledge Management (CIKM '22), October 17--21, 2022, Atlanta, GA, USA}
\acmPrice{15.00}
\acmDOI{10.1145/3511808.3557136}
\acmISBN{978-1-4503-9236-5/22/10}

\begin{document}


\title{BRIGHT - Graph Neural Networks in Real-Time Fraud Detection}


\author{Mingxuan Lu}

\affiliation{%
\institution{Shanghai Jiao Tong University}
\city{Shanghai}
\country{China}
}
\email{mingxuan.lu@sjtu.edu.cn}

\author{Zhichao Han}
\affiliation{%
\institution{eBay Inc.}
\city{Shanghai}
\country{China}
}
\email{zhihan@ebay.com}

\author{Susie Xi Rao}
\affiliation{%
  \institution{Swiss Federal Institute of Technology in Zurich (ETHZ)}
  \city{Zurich}
  \country{Switzerland}
}
\email{raox@inf.ethz.ch}

\author{Zitao Zhang}
\affiliation{%
\institution{eBay Inc.}
\city{Shanghai}
\country{China}
}
\email{zitzhang@ebay.com}

\author{Yang Zhao}
\affiliation{%
\institution{eBay Inc.}
\city{Shanghai}
\country{China}
}
\email{yzhao5@ebay.com}

\author{Yinan Shan}
\affiliation{%
\institution{eBay Inc.}
\city{Shanghai}
\country{China}
}
\email{yshan@ebay.com}

\author{Ramesh Raghunathan}
\affiliation{%
\institution{eBay Inc.}
\city{Shanghai}
\country{China}
}
\email{raraghunathan@ebay.com}

\author{Ce Zhang}
\affiliation{%
  \institution{Swiss Federal Institute of Technology in Zurich (ETHZ)}
\city{Zurich}
 \country{Switzerland}
}
\email{ce.zhang@inf.ethz.ch}

\author{Jiawei Jiang}
\affiliation{%
  \institution{School of Computer Science, \\ Wuhan University}
  \city{Wuhan}
  \country{China}
}
\email{jiawei.jiang@whu.edu.cn}

\renewcommand{\shortauthors}{Lu, et al.}

\begin{abstract}
Detecting fraudulent transactions is an essential component to control risk in e-commerce marketplaces. Apart from rule-based and machine learning filters that are already deployed in production, we want to enable efficient real-time inference with graph neural networks (GNNs), which is useful to catch multihop risk propagation in a transaction graph. However, two challenges arise in the implementation of GNNs in production.  
First, future information in a dynamic graph should not be considered in message passing to predict the past. 
Second, the latency of graph query and GNN model inference is usually up to hundreds of milliseconds, which is costly for some critical online services.
To tackle these challenges, we propose a Batch and Real-time Inception GrapH Topology (BRIGHT) framework to conduct an end-to-end GNN learning that allows efficient online real-time inference.

BRIGHT framework consists of a graph transformation module (Two-Stage Directed Graph) and a corresponding GNN architecture (Lambda Neural Network). 
The Two-Stage Directed Graph guarantees that the information passed through neighbors is only from the historical payment transactions. It consists of two subgraphs representing historical relationships and real-time links, respectively.
The Lambda Neural Network decouples inference into two stages: batch inference of entity embeddings and real-time inference of transaction prediction.
Our experiments show that BRIGHT outperforms the baseline models by >2\% in average w.r.t.~precision. Furthermore, BRIGHT is computationally efficient for real-time fraud detection. 
Regarding end-to-end performance (including neighbor query and inference), BRIGHT can reduce the P99 latency by >75\%.
For the inference stage, our speedup is on average 7.8$\times$ compared to the traditional GNN.


\end{abstract}

\ccsdesc[500]{Computing methodologies~Neural networks}
\ccsdesc[500]{Information systems~Enterprise applications}
\ccsdesc[300]{Theory of computation~Dynamic graph algorithms}

\keywords{graph neural networks, fraud detection, heterogeneous graph, dynamic graph, graph inference}


\maketitle

\section{Introduction}
Fraudulent transaction is one of the most serious threats to online financial services
today. This problem is exacerbated by the growing sophistication of business transactions using online payment applications (e.g., eBay and Alipay) and payment cards \cite{LAUNDERS2013150,Wang_2021}. Fraudsters use a range of tactics, including paying with stolen credit cards,
chargeback fraud involving complicit cardholders, selling fake e-gift cards, or creating schemes to create fraud ring attacks against multiple merchants\footnote{\url{https://www.verifi.com/in-the-news/need-know-fraud-rings/} (last accessed: April 22, 2022).}. In this work, our goal is to detect risky transactions on a real-world e-commerce platform.

In e-commerce marketplaces, unauthenticated transactions are a major part of buyer risk. 
In our previous generation of detection engine,
it is observed that the {\em entities} linked to transactions, such as shipping addresses and device machine IDs, are key clues in detecting 
frauds. Based on these entities, hundreds of patterns are summarized as features in machine learning models or as rules in decision engines.
However, the feature engineering of these linkage patterns, which are designed by human experts, is only feasible within 1-hop horizon in the graph.
Beyond 1-hop, it is still quite inefficient for human experts to explore and interpret, due to the huge amount of derived features generated by aggregating original features along risk propagation paths.

\textbf{Characteristics of Transaction Graphs}. Two vital characteristics of fraudulent transactions have been raised \cite{rao2020suspicious} in a similar application of massive suspicious registration detection. First, transactions and relevant entities naturally form a graph with \ul{heterogeneous nodes} (e.g., IP address, email, accounts). Those fraudulent transactions tend to share certain common risky entities, such as the device and IP address. Second, fraud detection in transaction graphs should consider the problem of \ul{temporal dynamic} because the accounts used by fraudsters and legitimate users usually generate distinctive activity events in separate time periods.

\textbf{Challenge 1: Prevention of learning from future transactions}.
{\em In a temporally dynamic graph, can we use information flowing from vertices whose timestamps are different from the current vertex?}
For the problem of fraud detection, claim, chargeback, and suspension histories are usually considered top-ranking features in fraud detection models, in which timestamp is an important property.
For a specific time, if timestamp is considered, there are two categories of events, historical and future events.
Intuitively, the feature patterns of the upcoming events linked by entities may allow models to foresee the risk. However, this assumption is impractical in real-time applications, as future vertices have not appeared in the graph yet. 
Therefore, in this work, we strictly constrain the model to utilize only the past transaction.

\textbf{Challenge 2: Efficiency of Graph Inference}.
During graph inference, different financial applications entail diverse latency requirements.
Some use cases, such as Know-Your-Customer (KYC) procedures during account registration and post-transaction evaluation for anti-money laundry detection, may tolerate relatively high latency.
However, detecting fraud transactions requires a quick response, since end users are often sensitive to latency. 
Generally, graph neural network (GNN) models aggregate information from neighbors within at least 2-hops in the graph; therefore,
each inference typically takes hundreds of milliseconds or even several seconds, which cannot meet internal requirements and user-side expectation on system latency.

To address the aforementioned challenges, we propose the BRIGHT (Batch and Real-time Inception GrapH Topology) framework.

\begin{enumerate}
    \item We propose a novel graph transformation method for temporally dynamic transaction graphs, yielding a Two-Stage Directed (TD) Graph storage strategy.
    This construction is time-sensitive, and
    can be utilized to perform time-aware message passing in real-time.
    
    \item Based on the transformed TD graph, we introduce a Lambda\footnote{Lambda architecture is a hybrid data processing approach that handles a large amount of data both in batch and in streaming fashion. For an overview on Lambda architecture in data processing, we refer readers to \url{https://en.wikipedia.org/wiki/Lambda_architecture} (last accessed: April 22, 2022).} Neural Network (LNN) architecture that supports both batch processing and real-time processing.
    LNN leverages snapshot aggregation and avoids foreseeing future information during training. We have also improved the efficiency of real-time inference using the Lambda architecture.
    
    \item Our experiments show that BRIGHT outperforms the baseline models by more than 2\% w.r.t.~average precision. Additionally, it is computationally efficient for real-time fraud detection. BRIGHT reduces more than 75\% P99\footnote{The P99 (99th percentile) latency is the worst latency that was observed by 99\% of all requests.} end-to-end inference latency (including neighbor query and graph inference), compared to traditional GNN inference frameworks. 
    For the inference stage, our speedup is on average 7.8$\times$ over the traditional GNN.
\end{enumerate}

\section{Preliminaries}
\label{sec:priliminary}

In our previous works, DHGReg \cite{rao2020suspicious} and DyHGN \cite{rao2022attention}, we have studied the dynamics of graph modeling in account and transaction networks.\footnote{We have discussed fraud detection with heterogeneous GNNs in an industrial billion-scale dataset, and conducted human evaluations against GNN predictions in xFraud \cite{rao2021xfraud}. There, we did not include the discussion of temporal aspects.} Specifically, DHGReg \cite{rao2020suspicious} is designed to detect massive suspicious registrations through a dynamic heterogeneous graph neural network. DHGReg is composed of two types of subgraphs: (1) structural subgraphs in each timestamp that reflect the links between different types of entity; and (2) a temporal subgraph that links the structural subgraphs via temporal edges.  As such, the time dimensions are unrolled in a large graph together with structural edges. With this graph construction technique, DHGReg outperforms simple baselines such as GCN and GAT in capturing suspicious registrations on time.

DyHGN \cite{rao2022attention} is built on top of DHGReg \cite{rao2022attention} and has studied to incorporate diachronic temporal embeddings \cite{goel2020diachronic} while still keeping structural subgraphs. It also explores the replacement of simple convolution layers with more complex ones, such as a heterogeneous transformer layer \cite{hu2020heterogeneous}. We also discuss graph structure and data distribution over time, as well as their influence on GNN modeling dynamics.

However, we notice three issues when deploying these prototype models in production:

\begin{itemize}
    \item \textbf{(I1)} The bi-graph structure (transactions and entities) tends to deplete GPU memory when the graph scales up to millions of vertices and edges;
    
    \item \textbf{(I2)} There is future information leakage in the training of predictive models. In the GNN message passing, features or embeddings from future events are also included when evaluating the risk of past events.
    
    \item \textbf{(I3)} The latency of the neighbor query is more than hundreds of milliseconds or even seconds, which is not ideal for a real-time responsive system. 
\end{itemize}

\section{Research Question and Methodology}

In this work, we propose BRIGHT (Two-Stage Directed Graph transformation + Lambda Neural Network) to tackle these remaining issues when deploying GNN in production. 
\begin{itemize}
    \item To solve \textbf{(I1)}, we keep the structural subgraphs of DHGReg and partition the graphs to support both batch and online learning. 
    
    \item To solve \textbf{(I2)}, we strictly control that only historical reference transaction information is used to predict target transactions. 
    
    \item To reduce the overhead of querying neighbors in \textbf{(I3)}, we enable batch inference and store entity embeddings in a key-value store.
    These entity embeddings are then used in online inference to decrease inference latency. 
\end{itemize}

\begin{figure*}[t!]
    \centering
    \includegraphics[width=0.81\textwidth]{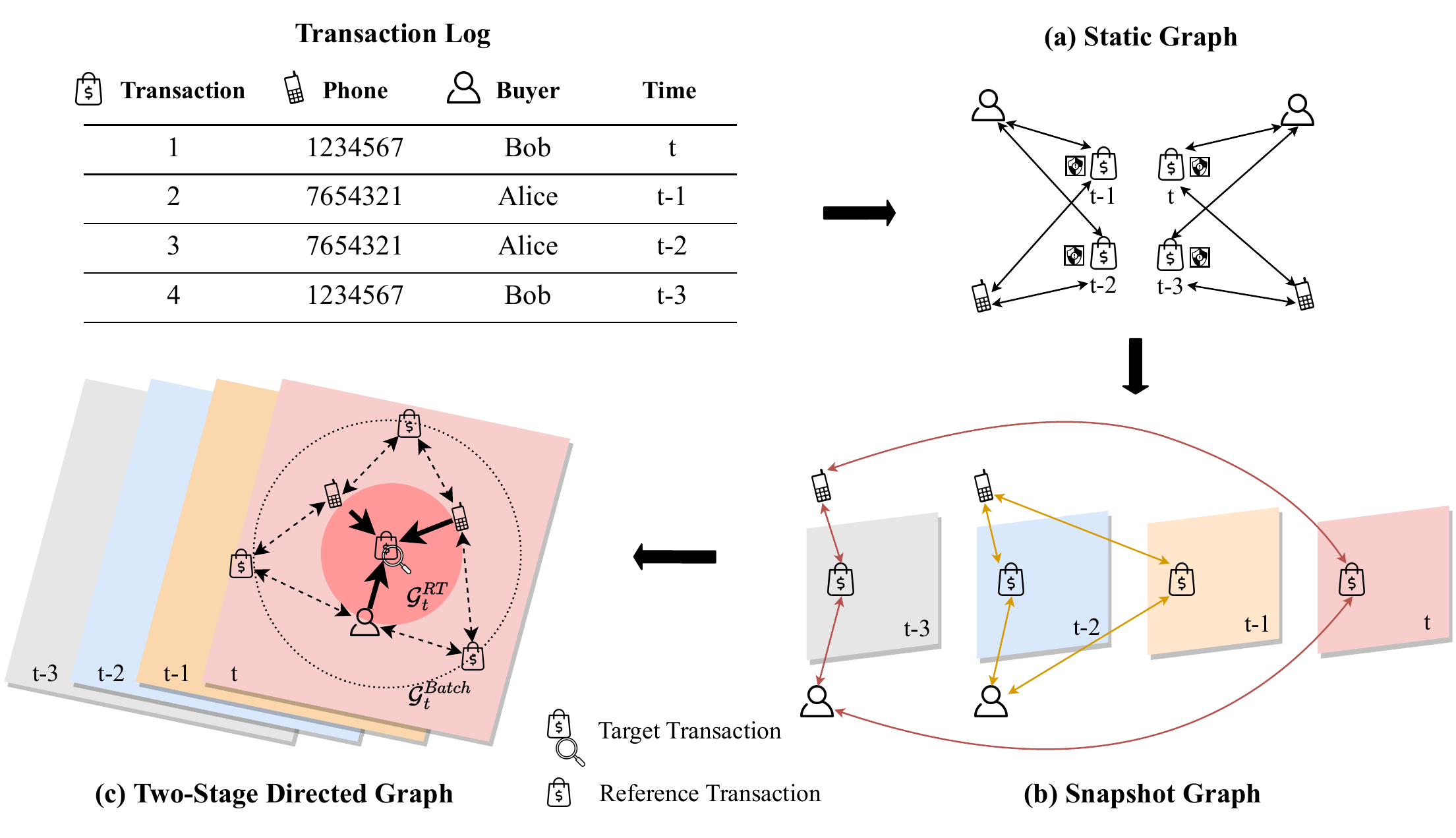}
    \vspace{-1em}
    \caption{Graph Transformation in BRIGHT. RT: Real-Time.}
    \label{fig:graph-trfm}
\end{figure*}

In this section, we introduce the research questions we want to answer with BRIGHT (Sec.~\ref{sec:research-q}), illustrate the graph transformation module (Sec.~\ref{sec:td-graph-construct}), and explain the Lambda Neural Network (Sec.~\ref{sec:lnn}). 

\subsection{Research Question}\label{sec:research-q}

With BRIGHT, our goal is to answer these two research questions.

\begin{itemize}
    \item \textbf{(Q1)} How could we construct a dynamic graph that allows us to control the direction of the information flow and effectively support online inference? 

    \item \textbf{(Q2)} How could we design a graph neural network architecture that is efficient for online inference?
\end{itemize}

\subsection{Time-sensitive Graph Construction} \label{sec:td-graph-construct}

In this work, transaction fraud detection is treated as a binary classification problem in an inductive setting on a dynamic heterogeneous graph. In a static transaction graph $\mathcal{G}$ (Fig.~ \ref{fig:graph-trfm} (a)), a vertex $v \in \mathcal{V}$ has a type $\tau(v) \in \mathcal{A}$, 
where $ \mathcal{A} := \{ transaction, entity \} $. An edge $e \in \mathcal{E}$ links from a $transaction$ vertex ($txn$ for short hereafter) to an $entity$ vertex.

\begin{table}[t!]
\centering
\caption{Notations.}
\label{tab:notations}
\vspace{-1em}
\begin{tabular}{ll}
\toprule
\textbf{Notation}        & \textbf{Description}                  \\ \midrule
$ \mathcal{G} $ & Static graph   \\
$ \mathcal{V} $ & Vertices on the static graph \\
$ \mathcal{E} $ & Edges on static graph \\ 
$v$ & Transaction or entity on the \\ 
$ $ & static graph \\
$e$ & Transaction-entity linkage on the \\
$ $ & static graph \\
$txn$ & Transaction vertex on the static graph \\ 
$entity$ & Entity vertex on the static graph \\
$\mathcal{T}$ & Timestamp set/Time window in a partition, \\
& $ \mathcal{T} := \{0,1,...,t\} $ \\
$ \mathcal{G_T} $ & Snapshot graph where \\ 
$ $ & each vertex has a timestamp \\
$ \mathcal{G}^{RT}_{T}$ & Real-Time (RT) Graph, used in real-time inference \\[3pt]
$ \mathcal{G}^{Batch}_{T}$ & Batch Graph for batch inference \\[3pt]
$ \mathcal{G}^{TD}_{T}$ & Two-Stage Directed Graph, \\
$ $ & union of $\mathcal{G}^{RT}_{T}$ and $\mathcal{G}^{Batch}_{T}$ \\
$ t $ & Time period snapshot $t$, a time duration (day) \\ 
$ \mathcal{G}^{RT}_{t}$ & Subgraph of Real-Time (RT) Graph on $partition_t$ \\[3pt]
$ \mathcal{G}^{Batch}_{t}$ & Subgraph of Batch Graph on $partition_t$ \\[3pt]
$snapshot_t$ & Time period snapshot $t \in \mathcal{T}$ \\
$partition_t$ & Time period partition $t \in \mathcal{T}$, \\
              & which is the partition of target transaction \\
              & at time $t$ that also contains reference transaction \\  
              & from historical time snapshots \\

$X_t$ & Feature embeddings of transactions on $partition_t$ \\
$txn_t$ & Transaction on $snapshot_t$ \\
$entity_t$ & Entity nodes on $snapshot_t$  \\
$txn^{tgt}_t$ & Target transaction on $partition_t$  \\
$txn^{ref}_t$ & Historical reference transaction on $partition_t$ \\
$entity^{tgt}_t$ & Entity nodes on $partition_t$  \\

\bottomrule
\end{tabular}
\end{table}

Nodes $txn$ with unauthenticated chargeback claims from the customer system are marked as $1$, which are considered fraudulent transactions. The others are marked as $0$, which represents legitimate transactions. These labels are used in our binary classification task. These $txn$ with labels are called \textbf{target} transactions in our setting. The historical transactions that share the same entities $entity$ used in the target transaction, such as emails and shipping addresses, we call them \textbf{reference} transactions, which do not have any labels in our experiments.

To construct a Two-Stage Directed Graph (TD Graph) to support Lambda Neural Network (LNN), our graph construction consists of the following steps, as illustrated in Fig.~\ref{fig:graph-trfm}.

\begin{itemize}
    \item \textbf{(a) Static Graph.} The static graph is constructed from months of transaction logs by transforming them into a bi-graph (transaction-entity graph). Note that the edges here are bidirectional. 
    \item \textbf{(b) Snapshot Graph.} Each node in the static graph is placed in its corresponding timestamp snapshot. Reference transactions are linked to entities that are on the same timestamp snapshot as the corresponding target transactions.
    \item \textbf{(c) Two-Stage Directed Graph.} A TD Graph is stored in target transaction partitions. A partition is made up of several timestamps that are the temporal snapshots in which all related transactions and entities are. A TD graph is equivalent to not only a snapshot graph, but also a simplified topology view for target transactions in the partition. The TD Graph is used for LNN learning and inference.
\end{itemize}

We now discuss in detail how the three types of graphs are constructed in Fig.~\ref{fig:graph-trfm}.

\subsubsection{\textbf{Static Graph Construction}}

To collect neighbor features to assess the risks of transaction fraud, multiple entities used in the checkout sessions are considered neighbors of $txn$ nodes. These entities, including shipping address, E-mail, IP address, device ID, contact phone, payment token, and user account, are represented as $entity$ nodes in $\mathcal{G}$.

Each $txn$ vertex represents a checkout transaction along with a unique transaction ID, 
linked with multiple $entity$ vertices such as shipping addresses, E-mails, contact phones that buyers need to confirm on checkout pages. 
Most of the $entity$ vertices are also linked to multiple $txn$ vertices. 

Given a set of target transactions, a static graph $\mathcal{G}$ can be constructed from their records. A transaction record could be broken down into a $txn$ vertex and several $entity$ vertices. Edges $e$ are placed between entities and transactions that use these entities. Reference transactions within a given observation window are extracted if they share the same entities. Edges are also added between reference transactions and common entities.

\subsubsection{\textbf{Snapshot Graph Construction}}

A time snapshot $t \in \mathcal{T}$, where $ \mathcal{T} := \{0,1,...,t\} $, represents a period of time duration, e.g., hour or day. $\mathcal{T}$ is the time window in one Snapshot Graph. In our experiments, the time period for each snapshot is one day.
A snapshot vertex $v_t \in \mathcal{V_T}$ represents the static vertex. The snapshot vertex $v_t$ is a transformation of $v$ at snapshot $t$. The workflow to construct a snapshot graph is described below.

\begin{enumerate}
    \item For each $txn$ node on the static graph $\mathcal{G}$, we construct $txn_t$ on $snapshot_t$, which is the time period in which the transaction was created.
    \item For each $entity$ node directly linked to the transaction nodes $txn_t$, we create the node $entity_t$ on $snapshot_t$, sharing the same transaction creation time. There may be multiple instances of the same entity in the snapshot graph $\mathcal{G_T}$ if it is linked to transactions created in various snapshots.
    \item Create edges between the target transaction nodes $txn_t$ and the 1-hop entity neighbor $entity_t$.
    \item Create edges between the reference transaction nodes $txn_i$ and 1-hop entity neighbor $entity_t$,
    where $0 \leq i \leq t$.
\end{enumerate}

On Snapshot Graph $\mathcal{G}_T$, a transaction node may have two roles. One is the target transaction; the other is the reference transaction if it shares the same entity as another target transaction. In Fig. \ref{fig:graph-trfm} (b), the transaction in snapshot $t-3$ is a reference transaction to the transaction in snapshot $t$. Meanwhile, it can also be a target transaction in snapshot $t-3$ when we predict its label.

Potential roles for a single $txn_t$ node make it difficult to isolate future information from message passing if there is no further filtering or sampling based on transaction timestamps. Some edges can be in $\mathcal{G}^{RT}_T$ and $\mathcal{G}^{Batch}_T$ (which we will introduce in Sec.~\ref{sec:two-stage}), which leads to future information leakage. We want to avoid neighbor sampling due to the high latency in existing SAGE neighbor sampling approaches, as shown in \cite{zhang2021graph}. This brings us to our design of Two-Stage Directed Graph shown in the next section.

\subsubsection{\textbf{Two-Stage Directed Graph}}
\label{sec:two-stage}

Two-Stage Directed (TD) graph simplifies the topology view from the target transaction side, making it easy to split the graph into Real-Time Graph $\mathcal{G}^{RT}_T$ and Batch Graph $\mathcal{G}^{Batch}_T$. We show in Fig.~\ref{fig:graph-trfm} (c) these two parts of the graphs that are taken later by LNN as input for various parts of the network. $\mathcal{G}^{RT}_T$ is the subgraph with thick directed edges (the dark red circle), while $\mathcal{G}^{Batch}_T$ is the subgraph with bidirectional edges (the light red circle).

\begin{figure}[!t]
    \centering
    \includegraphics[width=0.5\textwidth]{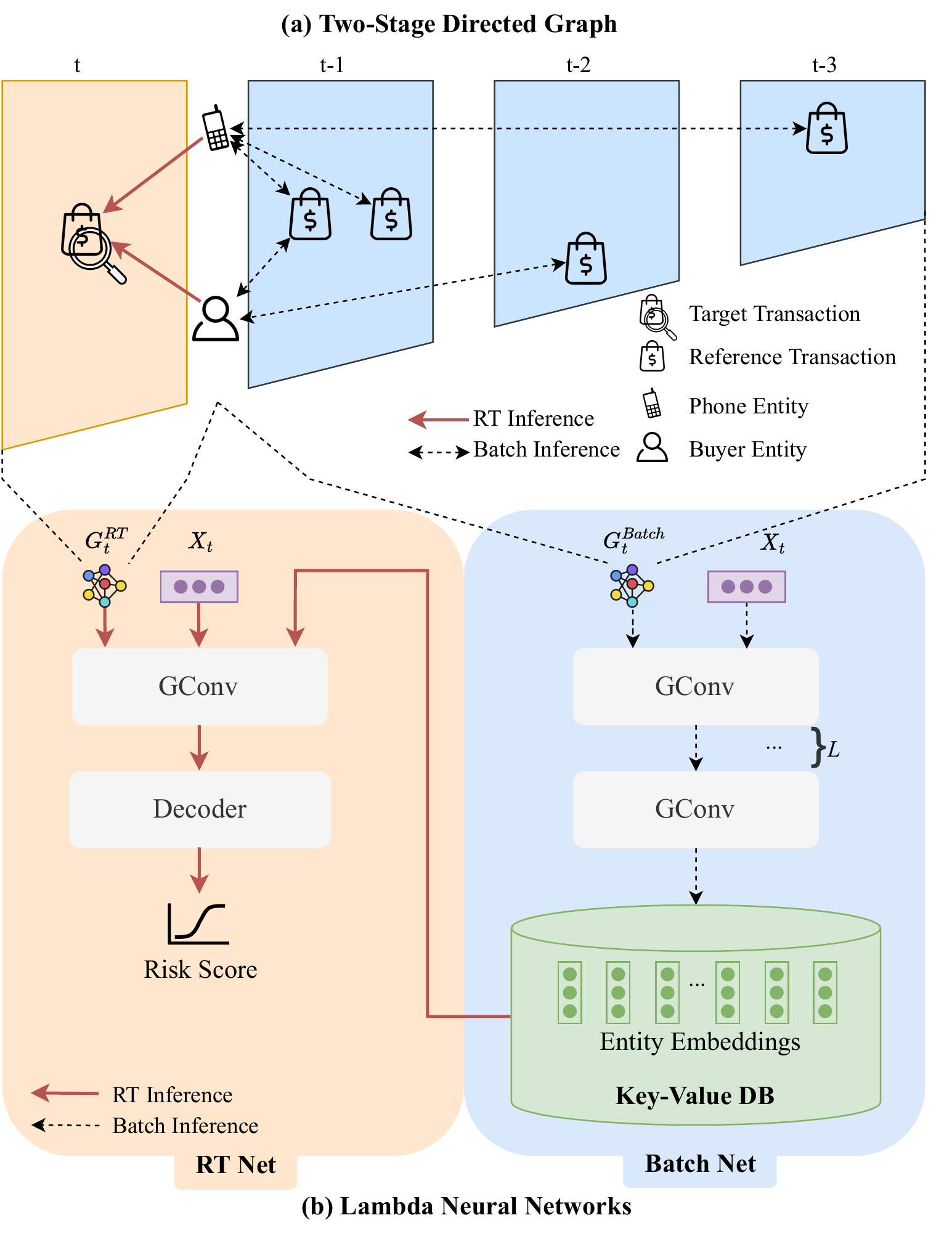}
    \vspace{-1em}
    \caption{Two-Stage Directed Graph (a) and Lambda Neural Network (b). RT: Real-Time, GConv: Graph convolution.}
    \label{fig:graph-dds-n-lambda}
\end{figure}

The partition of target transaction is introduced to isolate transaction roles, because a transaction can act as both a target/reference transaction, as we discuss above. All reference transaction nodes sharing common entities are stored with the target transactions in the same partition. On the TD graph, there is only one role for transactions, either a target transaction or a reference transaction. There are multiple snapshots $snapshot_i$ in one target partition $partition_t$, $i \leq t$, as illustrated in Fig.~\ref{fig:graph-dds-n-lambda} (a). The workflow to construct a TD graph is described below.

\begin{enumerate}
    \item Place target transactions $txn_t$, their corresponding 1-hop entities $entity_t$, and the reference     
          transactions $txn_i$, where $0 \leq i \leq t$ into $partition_t$. 
          These transactions are represented as $txn^{tgt}_t$ and $txn^{ref}_t$. 
          The entities are represented as $entity^{tgt}_t$.
          Reference transactions may be put into multiple partitions if their related target transactions are in different time snapshots.
    \item Link $txn^{ref}_t$ and $entity^{tgt}_t$ with bi-directed edges, 
          which forms the Batch Graph $\mathcal{G}^{Batch}_{t}$. It is used for batch inference of entity representations.
    \item Link $entity^{tgt}_t$ and $txn^{tgt}_t$ with directed edges from the entity to the target transaction, 
          which forms the Real-Time Graph $\mathcal{G}^{RT}_{t}$. It is used for real-time inference of transaction risk.
    
\end{enumerate}

The edges on Batch Graph are bi-directed, which enables bi-directional inference with deep GNN models on Batch Graph. The edge types for the TD Graph are represented in Tab.~ \ref{tab:dds-et}. The edges on Real-Time Graph are uni-directed, which are always from entities to the target transactions. The number of layers in Real-Time Net is set to 1 for inference efficiency, so it is not necessary to make the edges bi-directed. Furthermore, with this setting, we ensure that no message is passed between target transactions across multiple timestamps even if they are on the same target partition.

Note that the design of the TD graph is in line with the Lambda Neural Network, which enables batch and online inference in the same framework. We elaborate the design of the LNN architecture in the following Sec.~\ref{sec:lnn}.
In Fig.~\ref{fig:graph-dds-n-lambda} (a), we show an illustrative example of the risk assessment of the transaction on $txn^{tgt}_t$ using TD Graph. 
The features of historical reference transactions $txn^{ref}_t$ are sent to $entity^{tgt}_t$ (phone and buyer in this example). They are the entity nodes 1-hop away from the target transaction. After bi-directed message-passing through the dotted edges, the embeddings (for phone and buyer in this example) of 1-hop neighbors are obtained. Then these embeddings are sent to the target transaction $txn^{tgt}_t$ through the directed edges (in red), which forms the graph used in real-time inference. With transformation into a TD Graph, the messages passed only from historical entities/transactions prior to $txn^{tgt}_t$ are guaranteed, which addresses the research question \textbf{(Q1)}.
\begin{table}[!t]
\centering
\caption{Two-Stage Directed Graph Edge Types.}
\label{tab:dds-et}
\vspace{-1em}

\begin{tabular}{ll}

\toprule
\textbf{Edge Type}  & \textbf{Description}     \\ 

\midrule

$txn^{ref}_{t} \leftrightarrow entity^{tgt}_{t}$ & Bi-directed edges between reference \\ 
$ $ & transactions and entities on  \\ 
$ $ & the Batch Graph $\mathcal{G}^{Batch}_t$  \\ 

$entity^{tgt}_{t} \rightarrow txn^{tgt}_t$ & Directed edges from entities to target \\
$ $ & transaction nodes on \\ 
$ $ & the Real-Time Graph $\mathcal{G}^{RT}_t$ \\

\bottomrule
 
\end{tabular}
\end{table}

\subsection{Network Architecture}
\label{sec:lnn}

We propose a Lambda Neural Network (LNN) architecture, which treats the last hop edges in Real-Time Graph differently from those edges in Batch Graph from TD. This is illustrated in Fig.~\ref{fig:graph-dds-n-lambda} (b).

Lambda Architecture is a hybrid data processing approach that handles a large amount of data in both batch and streaming fashion. One of the benefits from Lambda Architecture is the balance between the data processing scalability and the data access latency. LNN architecture handles message passing from Batch Graph and Real-Time Graph in separate GNN blocks. Message passing on Batch Graph is conducted in batch jobs through Batch Net. Message passing on the Real-Time Graph is conducted in real-time inference through the Real-Time (RT) Net, deployed in online services.

Similar to the two-``tower" model architecture \cite{covington2016deep, 48840googletwotower, yang2020mixed, wang2021exploring, wang2021cross, yu2021dual} (see Sec.~\ref{sec:related}), we design LNN to support two stages of processing (batch and real-time inference). The LNN architecture, together with the TD graph, addresses the research question \textbf{(Q2)}, which is our proposed solution for online inference in a dynamic setting.

As illustrated in Fig.~ \ref{fig:graph-dds-n-lambda}, LNN consists of a Batch Net (in blue) and a Real-Time (RT) Net (in orange).  Batch Net is a stack of GNN layers, similar to DeepGCNs \cite{li2019deepgcns}. RT Net consists of a graph convolutional layer and a decoder, which could simply be a fully connected linear layer. The two stages are divided at the neighboring entities in one hop to $entity^{tgt}_{t}$ (phone and buyer in this example). The first stage is Batch Net inference, after which we obtain the entity embeddings learned from historical transactions and flush them into a key-value store. In the second stage, the RT Net fetches the entity embeddings from the key-value store, also takes the raw features from the target transactions, and then computes the transaction risk scores.

\subsubsection{\textbf{End-to-end Training}}

In the training phase, both the Batch Graph and the RT Graph are used for end-to-end LNN training. 
In each mini-batch, we sample one partition for the time window $\mathcal{T}$, where $ \mathcal{T} := \{0,1,...,t\} $.
The timestamp of each target transaction is within the partition time window. 
All timestamps for reference transactions are prior to those for target transactions.

Similar to traditional GNNs,
the inputs of Batch Net include the features of reference transactions $X_t$ and Batch Graph $G^{Batch}_t$, as illustrated in Eq.~\ref{eqn:batch-infer}:

\begin{equation} \label{eqn:batch-infer}
h^{Batch}_t = BatchNet(X_t, G^{Batch}_t)
\end{equation}

Each graph convolutional layer of Batch Net is generic with a hidden state $h^l$ of the layer $l$, as illustrated in Eq.~\ref{eqn:batch-layer}. $\mathcal{N}(v)$ is all the neighbors of the vertex $v$, which is $txn^{ref}_t$ or $entity^{tgt}_t$. $\odot$ is a generalized node aggregation function. $\cup$ represents concatenation, $W$ weight matrices. 

\begin{equation} \label{eqn:batch-layer}
h^{l+1}_v = \sigma (
W \cdot \odot (
  \{h^l_v\} \cup \{h^l_u, \forall u \in \mathcal{N}(v)\}
)
) + h^l_v
\end{equation}



As input, RT Net takes features from target transactions and entity embeddings learned from Batch Net inference. The graph convolutional layer of RT Net shares the same form as that of Batch Net, as is illustrated in Eq.~\ref{eqn:rt-layer}. 
One difference is that there is only one convolutional layer in the RT Net for the sake of inference efficiency. \footnote{\textcolor{black}{In consideration of both inference efficiency and the prevention of data leakage, a single layer is chosen for RT Net. (1) Obviously, one convolutional layer provides lower inference latency. (2) Currently, only one-hop neighbours to the target transactions are linked by unidirectional edges, the other neighboring nodes are connected by bidirectional edges. Therefore, if there are more than one convolutional layers, we would have a data leakage problem, where the future information could be used to predict nodes in the past. }}
Furthermore, RT Net isolates the aggregation of messages between different target transaction nodes $txn^{tgt}_t$ within the same target snapshot partition. $\phi$ is a linear projection for the transaction features, which assures that the transaction features and the embeddings have the identical dimensions.

\begin{dmath} \label{eqn:rt-layer}
h^{RT}_{txn^{tgt}_t} = \sigma (
W \cdot \odot (
  \{\phi (X_{txn^{tgt}_t})\} \cup \{h^{Batch}_{entity^{tgt}_t}, \forall entity^{tgt}_t \in \mathcal{N}(txn^{tgt}_t)\}
)
) + \phi (X_{txn^{tgt}_t})
\end{dmath}


The risk  score for target transaction $txn^{tgt}_t$ is decoded from the real-time transaction embedding $h^{RT}_{txn^{tgt}_t}$ through a fully connected linear layer, which is illustrated in Eq.~\ref{eqn:dec}.

\begin{equation} \label{eqn:dec}
score_{txn^{tgt}_t} = W \cdot h^{RT}_{txn^{tgt}_t}
\end{equation}

\subsubsection{\textbf{Efficient Graph Inference}}

As described above, end-to-end learning uses the complete LNN architecture. 
When deployed in a production environment, LNN inference is decoupled into batch inference and real-time inference.

\textbf{(Batch Inference Stage.)}
For batch inference stage, the embeddings $h^{Batch}_t$ of entities $entity^{tgt}_{t}$ are periodically generated according to Eq.~\ref{eqn:batch-infer}. In our current design, the time period is one day. Finally, entity embeddings are generated and stored in a distributed key-value database for multiple downstream purposes. In our paper, these embeddings are used to detect fraud transactions. The embeddings are refreshed 
daily to serve real-time transaction risk evaluation.

\textbf{(Real-Time Inference Stage.)}
For real-time risk assessment, the second stage of LNN (namely RT Net) calculates the score by Eq.~\ref{eqn:dec}. 
The results are equivalent to end-to-end inference with the complete LNN adopted in the learning stage.
The inference latency is further significantly reduced due to the 1-layer RT Graph.
Since RT Graph only requires 1-hop neighbors, which are already available in the transaction request, a 
graph query is not necessary during real-time inference. \textcolor{black}{Graph query refers to fetching nodes and their features from a graph database such as Neo4j. Illustrative queries in Cypher\footnote{\url{https://neo4j.com/developer/cypher/}.} are given in Tab.~\ref{tab:cypher}. For a single layer RT Net, the number of hops is set to 2. It is possible to include more hops in the RT Net, as we have discussed in Sec.~\ref{sec:extended-graph}. In the experiments of multiple hops, we also use one convolution layer for RT Net, to refrain from data leakage.  }

\begin{table}[]
    \centering
    \caption{\textcolor{black}{Cypher Queries for Different Number of Hops (in the Extended Graph) in the RT Net. RT: Real-Time.}}
    \label{tab:cypher}
    \vspace{-1em}
    \resizebox{\linewidth}{!}{
    \begin{tabular}{c|c}
        \toprule
        \textbf{Number of Hops} & \textbf{Cypher Query} \\
        \midrule
        \textbf{2} & 
        \begin{tabular}[c]{@{}l@{}}\textbf{MATCH}\\ (target:transaction)-{[}{]}-\textgreater{}(:entity)-\textgreater{}{[}{]}-\textgreater{}(nbr2:transaction)\\ \textbf{WHERE} transaction=\$transaction\_id\\ \textbf{RETURN} nbr2\end{tabular}  \\
        \midrule
        \textbf{4} &  \begin{tabular}[c]{@{}l@{}}\textbf{MATCH}\\ (target:transaction) -{[}{]}-\textgreater{}(:entity)-\textgreater{}{[}{]}-\textgreater{}(nbr2:transaction)\\                                -{[}{]}-\textgreater{}(:entity)-\textgreater{}{[}{]}-\textgreater{}(nbr4:transaction)\\ \textbf{WHERE} transaction=\$transaction\_id\\ \textbf{RETURN} nbr2, nbr4\end{tabular} \\
        \midrule
        \textbf{6} & \begin{tabular}[c]{@{}l@{}}\textbf{MATCH}\\ (target:transaction)-{[}{]}-\textgreater{}(:entity)-\textgreater{}{[}{]}-\textgreater{}(nbr2:transaction)\\                     -{[}{]}-\textgreater{}(:entity)-\textgreater{}{[}{]}-\textgreater{}(nbr4:transaction)\\                     -{[}{]}-\textgreater{}(:entity)-\textgreater{}{[}{]}-\textgreater{}(nbr6:transaction)\\ \textbf{WHERE} transaction=\$transaction\_id\\ \textbf{RETURN} nbr2, nbr4, nbr6\end{tabular} \\
        \bottomrule
    \end{tabular}}
    
\end{table}

\begin{table}[!t]
\centering
\caption{Dataset Summary ("M": Million). *The fraud ratios are only reported on the sampled datasets. TD: Two-Stage.}
\label{tab:data-summary}
\vspace{-1em}
\begin{tabular}{ccccc}
\toprule
\textbf{Split}                   & \textbf{Graph}    & \textbf{\#Nodes} & \textbf{\#Edges} & \textbf{Fraud\%* }               \\ \midrule
\multirow{3}{*}{\textbf{Training}}                  & Static   & 56.7M   & 153.9M  & \multirow{3}{*}{2.70\%} \\
                          & Snapshot & 56.7M   & 274.5M  &                         \\
                          & TD       & 66.4M   & 332.0M  &                         \\ \midrule
\multirow{3}{*}{\textbf{Testing}}                   & Static   & 9.63M   & 18.42M  & \multirow{3}{*}{2.30\%}\\
                          & Snapshot & 9.63M   & 30.39M  &                         \\
                          & TD       & 10.28M  & 32.88M  &                         \\ \bottomrule
\end{tabular}
\end{table}

\begin{figure}[!t]
    \includegraphics[width=6.5cm]{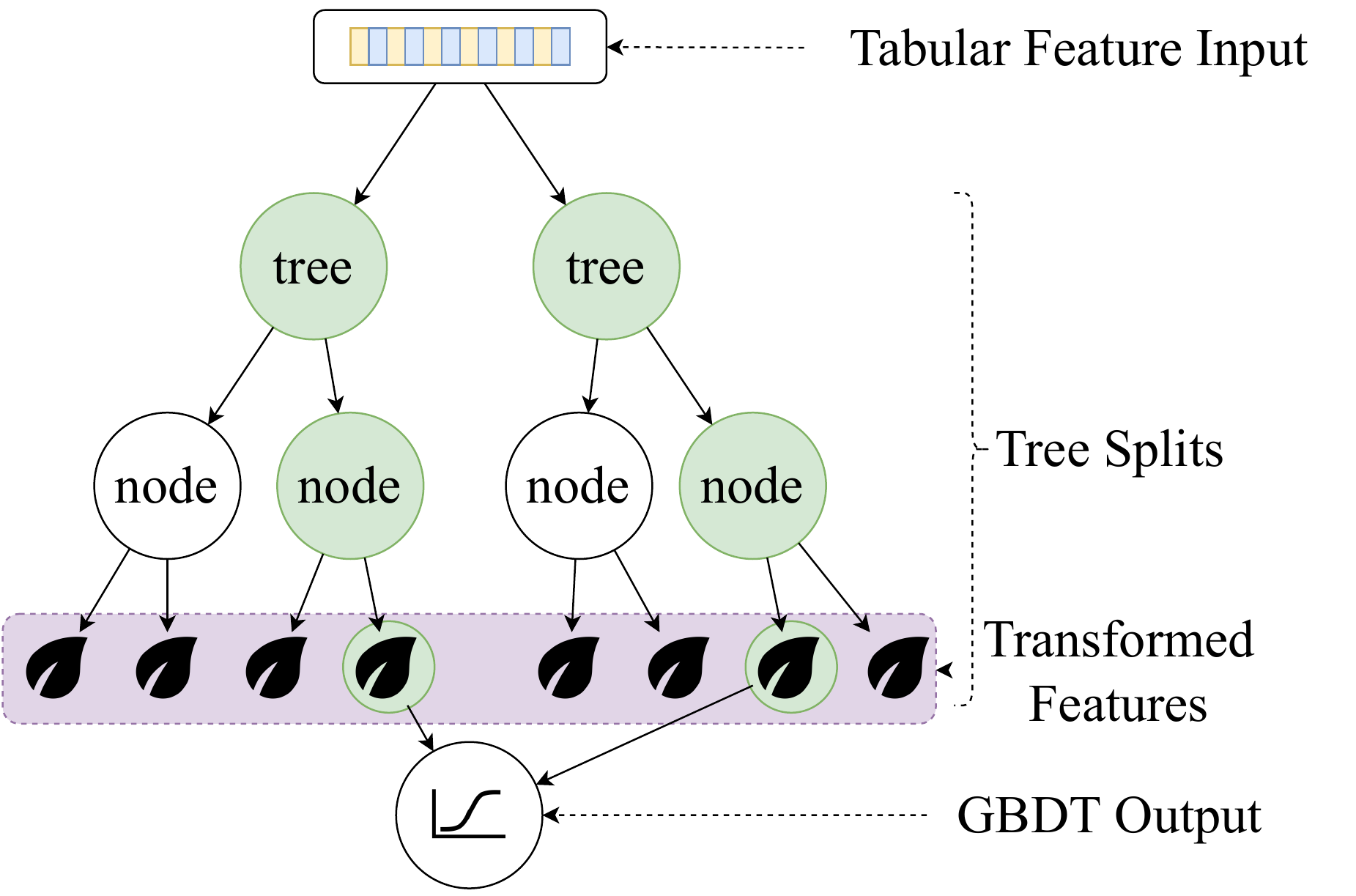}
    \vspace{-1em}
    \caption{GBDT as Feature Encoder.}
    \label{fig:gbdt-encoder}
\end{figure}

\section{Fraud detection experiment}

In this section, we perform an evaluation of BRIGHT's performance in the fraud detection task. 
\subsection{Dataset and Preprocessing} 
\label{sec:data}

We conduct experiments on real-world transaction records sampled from eBay marketplace to evaluate the proposed framework. 
Our dataset contains 7 million labeled transactions.\footnote{Note that in our experiments, only these 7 million transaction records have labels, the graphs in Tab. \ref{tab:data-summary} contain more transaction nodes yet without labels (reference transactions). It is common that not every transaction has a label, which could be due to delay in third-party service, as we elaborated in \cite{rao2021xfraud}.} Downsampling is conducted on legitimate transactions due to the label imbalance, as in \cite{rao2021xfraud}. The sampling ratio for fraud transactions is kept at ard.~ 2.5\%, which is empirically determined, as in other use cases at eBay. 
Entities in $\mathcal{G}$ include shipping addresses, device IDs, phones, payment tokens, IPs, and buyer accounts.

The data set is divided by the creation time of the target transactions into training and testing datasets. The first 80\% target transactions are used as a training dataset, and the last 20\% transactions are used for testing. The validation set for early stopping check is 10\%, randomly chosen from the training dataset. The statistics for the graph are illustrated in the Tab. \ref{tab:data-summary}. 

\subsection{Baselines}

LightGBM (LGB) \cite{ke2017lightgbm} is chosen as the baseline. 
There are two baseline models in our experiments built upon LightGBM. One is fully trained, namely, LGB (Bench), with 10k + trees, and was early-stopped after 32 iterations if the validation metrics have not improved. 
The other is a light version with 512 trees. The light model, namely the LGB (FE), is also utilized as a feature encoder (FE) for the LNN input.

\subsection{Feature Encoder}

Gradient Boosting Decision Tree (GBDT) models such as LightGBM \cite{ke2017lightgbm} dominate many tabular dataset applications \cite{gorishniy2021revisiting}, compared to existing popular deep model architectures designed specifically for tabular dataset, such as AutoInt \cite{song2019autoint} and TabNet \cite{arik2021tabnet}. For fraud detection on payment transactions, where most datasets are tabular, boosting models are widely used \cite{zhou2018fraud, cao2019titant}. For simplicity, a pre-trained  GBDT model is used as the feature encoder for tabular features. Similarly to Logistic Regression models on top of Decision Trees \cite{he2014practical}, the values of the chosen leaf nodes are used as encoded embeddings from the feature encoder, as illustrated in Fig. \ref{fig:gbdt-encoder}.

\subsection{Experiment Setup}

The transaction records sampled as described in Sec.~\ref{sec:data} are transformed into graphs using our method proposed in Sec.~\ref{sec:td-graph-construct}.

\textbf{(Data Loader.)}
The mini-batch strategy is used in GNN learning. 
In each mini-batch, there is only one target partition.
The maximum number of historical time snapshots ($\mathcal{T}$) in the target partition is set to 30. The size of the mini-batch depends on the merged community\footnote{Similar to ClusterGCN \cite{Chiang_2019clustergcn}, graph clustering structure is used in the BRIGHT training process. Historical neighboring reference transactions are 2 hops away from the target transactions. For the connected components, whose node sizes are larger than 32k, they are cut into smaller communities through Louvain \cite{Blondel_2008louvain}. Before training, small communities are merged in the same target partition. The node sizes of the merged communities are close to 32k. \textcolor{black}{The batch size is chosen by experience to balance disk IO and GPU memory. The impact of batch size in both efficiency and fraud detection performance could be investigated in our future work. Different batches are not overlapping in terms of target transactions that are predicted. But the features and nodes used for each prediction could be overlapping if they are used/linked in the neighborhood of the target transactions.}} node size, which is close to 32k. The transformed TD graph is used as input to the LNN model. 
$\mathcal{G}^{Batch}_t$ and $\mathcal{G}^{RT}_t$ are fed separately for LNN models (Batch Net and RT Net, respectively). 

\textbf{(Feature Encoding.)}
The features of $txn^{ref}_t$ and $txn^{tgt}_t$ are first encoded by LGB (FE). 
The embeddings $X^t$ generated by LGB (FE) are then fed to LNN, together with $\mathcal{G}^{Batch}_t$ and $\mathcal{G}^{RT}_t$. 
The features of $entity^{tgt}_t$ are set to zero. 
The vector size of the transaction embeddings and entity embeddings is 512 (equal to the number of trees). The remaining parameters to train GBDT are default values.

\textbf{(Model Architecture.)}
As in DeepGCN \cite{li2019deepgcns}, we choose GCN \cite{Kipf2017SemiSupervisedCW} and GAT \cite{velivckovic2017graph} for Batch Net, and GCN for RT Net.

\textbf{(Hyperparameters.)} We grid search for the number of GNN layers for Batch Net and the number of hidden units for both Batch Net and RT Net. 
The numbers of GNN layers are chosen from $\{4, 6, 8, 16\}$. 
The numbers of hidden units are chosen from $\{256, 512\}$. The learning rate is set to 0.001 for all experiments. 
The maximal epochs for training are 128. 
Regarding early stopping, we terminate the training if the loss over the validation dataset does not decrease within 16 epochs.
For each group of hyperparameters, we execute three runs and report the average results in Tab.~\ref{tab:exp-sim}.

\begin{table}[!t]
\centering
\caption{LNN Performances in Fraud Detection (Test Set).\\ \textcolor{black}{LGB: LightGBM, LNN: Lambda Neural Network.} }
\label{tab:exp-sim}
\vspace{-1em}
\begin{tabular}{ccc}
\toprule
\textbf{Model} & \textbf{ROC AUC} & \textbf{Average Precision} \\ \midrule
\textbf{LGB (FE)}     & 0.9255$\pm$0.0003   & 0.3984$\pm$0.0007             \\ 
\textbf{LGB (Bench)}  & 0.9247$\pm$0.0005   & 0.4201$\pm$0.0013             \\ 
\midrule
\textbf{\textcolor{black}{BRIGHT-}LNN (GAT)}  & \textbf{0.9286$\pm$0.0005}   & 0.4394$\pm$0.0009         \\ 
\textbf{\textcolor{black}{BRIGHT-}LNN (GCN)}  & 0.9276$\pm$0.0016   & \textbf{0.4420$\pm$0.0035} \\ \bottomrule
\end{tabular}
\end{table}

\begin{figure}[!t]
    \centering
    \includegraphics[width=7cm]{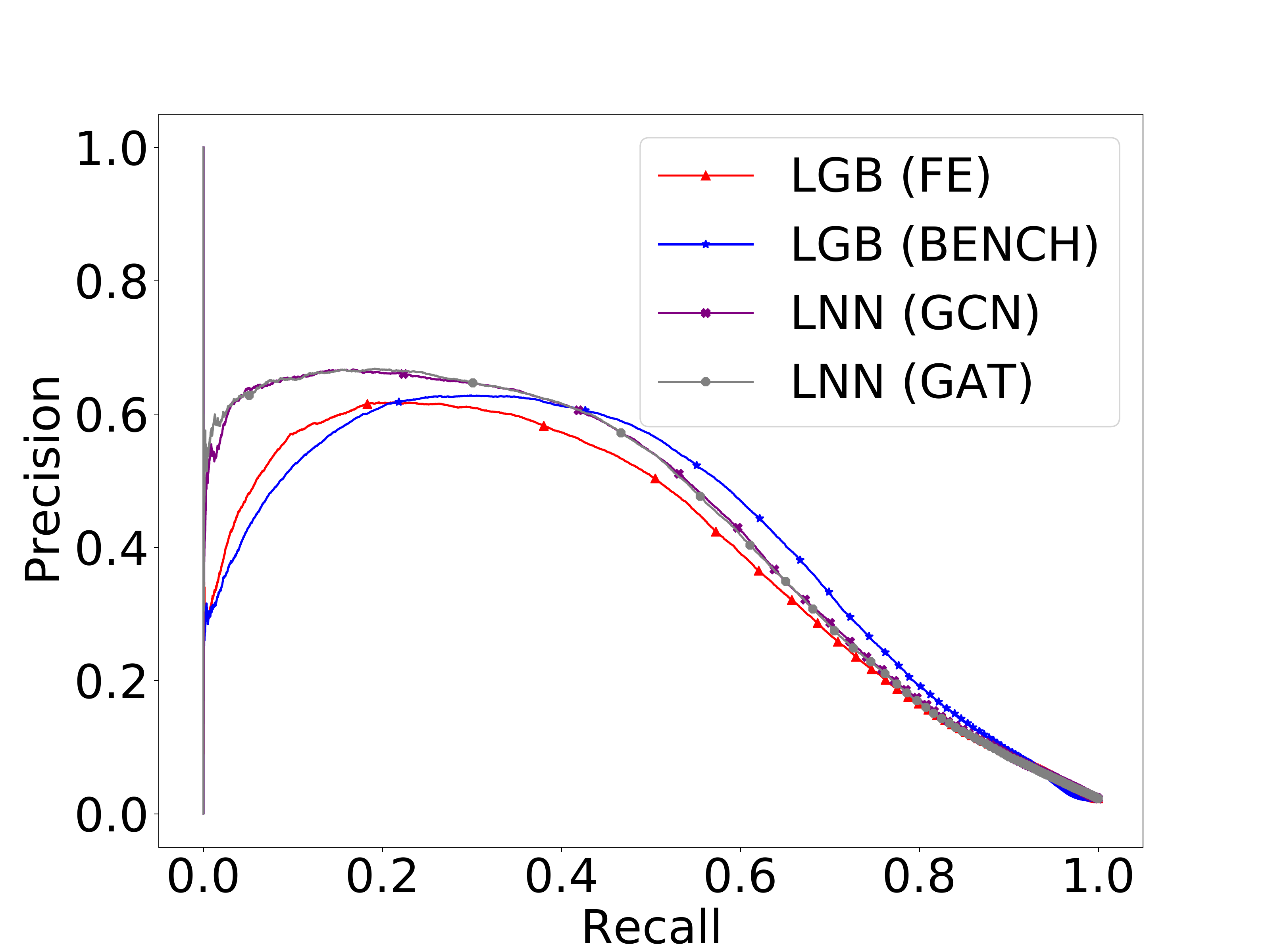}
    \vspace{-1em}
    \caption{Precision-Recall Curves.}
    \label{fig:auc-pr}
\end{figure}

\subsection{Quantitative Results on Fraud Detection}
We report the results of Average Precision (AP) and Area Under the Receiver Operating Characteristic Curve (ROC AUC) on the testing data in Tab.~ \ref{tab:exp-sim}. 
As can be seen, LNN outperforms LGB (FE) in terms of both ROC AUC and AP, by aggregating features through graph topology. LNN (GCN) has an average precision of 44.2\%.
Compared to LGB (Bench), which is the state-of-the-art model for tabular data, LNN (GCN) achieves an improvement of 2.2\%. 
We also plot the precision-recall curves in Fig.~\ref{fig:auc-pr}, and the curves show that LNN models can capture more fraud transactions with higher precision if we prioritize precision over recall. In our previous work \cite{rao2021xfraud}, we have discussed that in fraud detection tasks, we generally maintain a balance between precision and recall, while favoring precision over recall. This is because we try to reduce the workload of the business unit in examining false positives while trying to catch as many frauds as possible.

\section{Lambda Neural Network Architecture Efficiency}

Optimizing inference efficiency is another goal in our BRIGHT framework, which aims to minimize neighbor queries during GNN inference.
In this evaluation, we compare LGB (Bench), end-to-end LNN (GCN), and RT Net.

\subsection{Experiment Setup} 

For LNN (GCN) and RT Net, we use both CPU and GPU devices. In each test, a target transaction is evaluated and the system evaluates its risk of fraud.
We report the real-time inference latency, which is decoupled into three stages --
{\em graph query}, {\em feature collection}, and
{\em model inference}.


\textbf{(Graph Query.)} For GNN inference, neighboring reference transactions are required for graph construction and feature collection. Theoretically, key-value databases could be used to support query services for neighbor lookup. However, they are not good choices due to the complexity of the service implantation. Graph databases are natural choices as they provide a standard graph query language. In our experiments, Neo4j Community Edition\footnote{https://neo4j.com/} is used as our graph query service. 

We set up the database for one target partition, which is randomly selected from the test set. There are 0.7 million nodes and 3.8 million edges loaded into the graph DB. The default configuration is used for the Neo4j server. For LGB (Bench) and RT Net, there is no graph query test. LGB (Bench) does not need relational data for inference. In RT Net, the 1-hop entity IDs for embedding collection are in the request to predict the label of target transaction.

\textbf{(Feature Collection.)} Both raw features and entity embeddings are stored in a key-value database in our experiments. For simplicity, Redis\footnote{https://redis.io/} is chosen as our key-value database. For these 0.7 million records, we store their raw features and embeddings on the Redis server with the default configuration.

\textbf{(Model Inference.)} LGB (Bench) inference is conducted on CPU only. LNN (E2E) inference is an end-to-end full GNN inference, where we do not separate the Batch Net and RT Net. LNN (E2E) is equivalent to a traditional GNN inference workflow. The graph convolutional layers chosen for LNN is GCN. For the tests of RT Net inference, there is no Batch Net inference. Entity embeddings from batch inference are obtained from the key-value store. Both CUDA and CPU devices are tested for LNN and RT Net inference. Both models are implemented in PyTorch. For simplicity, there is no further optimization of the inference for the LNN models and RT Net models. We randomly choose 10k payment transactions for the latency test, recorded in milliseconds. 
We use one machine for the database server and inference client.
Regarding hardware, the machine is equipped with 32 Intel Xeon Gold 6230 CPUs, 64 GB memory, and one Nvidia Tesla V100 GPU.

\begin{table*}[t]
\caption{Inference Latency (recorded in millisecond). Numbers in brackets refer to  speedups compared to LNN (E2E). \textcolor{black}{In LNN (E2E) there is no separation of the Batch Net and RT Net. LGB: LightGBM, LNN: Lambda Neural Network, RT: Real-Time.}}
\label{tab:inference-latency}
\vspace{-1em}
\resizebox{\linewidth}{!}{
\begin{tabular}{c|cc|cc|cc|cc|cc}
\toprule
\textbf{Model}              & \multicolumn{2}{c|}{\textbf{LGB}} & \multicolumn{2}{c|}{\textbf{LNN (E2E, CUDA)}} & \multicolumn{2}{c|}{\textbf{LNN (E2E, CPU)}} & \multicolumn{2}{c|}{\textbf{\textcolor{black}{BRIGHT-}RT Net (CUDA)}}                                                                & \multicolumn{2}{c}{\textbf{\textcolor{black}{BRIGHT-}RT Net (CPU)}}          \\
\midrule
\textbf{Step}               
    & \textbf{Avg}   & \textbf{P99}   & \textbf{Avg}    & \textbf{P99}    & \textbf{Avg}    & \textbf{P99}    & \textbf{Avg}   & \textbf{P99}   & \textbf{Avg}   & \textbf{P99}   \\ \midrule
\textbf{Graph Query }       
    & -     & -     & 114.42 & 177.60 & 114.42 & 177.60 & -     & -     & -     & -     \\
\textbf{Feature Collection }
    & 0.06  & 0.13  & 0.08   & 0.20   & 0.08   
    & 0.20
    & 0.06 ({1.33$\times$})
    & 0.11 ({1.82$\times$})
    & 0.06 ({1.33$\times$}) 
    & 0.11 ({1.82$\times$}) 
    \\
\textbf{Feature Encoding}   
    & -     & -     
    & 51.88  & 72.16  & 51.88  & 72.16  
    & 51.70  (1.00$\times$)
    & 71.93  (1.00$\times$)
    & 51.70  (1.00$\times$)
    & 71.93  (1.00$\times$) \\
\textbf{Model Inference }   
    & 52.12 & 71.33 & 15.26  & 29.86  & 150.98 & 365.83 
    & 1.94  ({7.87$\times$})
    & 6.22  ({4.80$\times$})
    & 20.74 ({7.28$\times$}) 
    & 63.81 ({5.73$\times$})  \\ \midrule
\textbf{Total}              
    & 52.18 & 71.40 & 181.64 & 251.35 & 317.36 & 532.69 
    & 53.70  ({3.38$\times$})
    & 72.57  ({7.34$\times$})
    & 72.44  ({4.38$\times$})
    & 124.26 ({4.29$\times$}) \\ \bottomrule
\end{tabular}}
\end{table*}

\begin{table}[t!]
\caption{Model Inference Latency in Extended Graph (recorded in milliseconds). 
Numbers in brackets refer to speedups compared to LNN (E2E). \textcolor{black}{In LNN (E2E) there is no separation of the Batch Net and RT Net. LGB: LightGBM, LNN: Lambda Neural Network, RT: Real-Time.}}
\vspace{-1em}
\label{tab:inference-latency-khop}
\resizebox{\linewidth}{!}{
\begin{tabular}{c|c|cc|cc}
\toprule
\multicolumn{2}{c|}{\textbf{Model}} & \multicolumn{2}{c|}{\textbf{LNN (E2E)}} & \multicolumn{2}{c}{\textbf{\textcolor{black}{BRIGHT-}RT Net}}      \\ \midrule
\textbf{Device}        & \textbf{Hop}       & \textbf{Avg}        & \textbf{P99}        & \textbf{Avg}            & \textbf{P99}            \\ \midrule
\multirow{3}{*}{\textbf{CUDA}}           & 2         & 15.26      & 29.86      & 1.94 (7.87$\times$)   & 6.22 (4.80$\times$)   \\
              & 4         & 14.65      & 30.66      & 1.84 (7.96$\times$)   & 5.90 (5.20$\times$)   \\
              & 6         & 16.14      & 57.03      & 2.01 (8.01$\times$)   & 6.46 (8.83$\times$)   \\ \midrule
\multirow{3}{*}{\textbf{CPU}}            & 2         & 150.98     & 365.83     & 20.74 (7.28$\times$)  & 63.81 (5.73$\times$)  \\
              & 4         & 171.69     & 1044.73    & 16.21 (10.59$\times$) & 49.86 (20.95$\times$) \\
              & 6         & 216.36     & 1335.98    & 19.00 (11.39$\times$) & 58.44 (22.86$\times$) \\ \bottomrule
\end{tabular}}
\end{table}

\subsection{Experimental Results} 
As can be seen in Tab.~\ref{tab:inference-latency},
we compare LNN (E2E) and RT Net.
The end-to-end inference 99\% latency (99th percentile latency, a.k.a. P99) is significantly reduced from 251.35 ms to 72.57 ms using RT Net, that is, 71.13\% latency saving on CUDA devices. 
In CPU devices, the latency saving is even greater (76.67\%).
The reason for the performance improvement is that
the graph query and Batch Net inference are omitted in RT Net.
Compared to LGB, although RT Net introduces a more complex model architecture, only around 1.6\% overhead of P99 inference latency is introduced in the inference stage. 
Note that our latency numbers (<100ms) resonate the results reported by previous benchmarking papers in the GNN inference optimization \cite{zhang2021graph, zhou2021accelerating}.


\subsection{Discussion on Extended Graph Inference}
\label{sec:extended-graph}

In our previous experiments, we only choose two-hop neighbors (i.e., entities and their reference transactions) for the target transaction.
As readers might suspect, {\em what if neighbors of more hops are considered?}
To answer this question, we perform experiments to evaluate inference latency with 4-hop and 6-hop neighbors. We take a daily partition with up to 6-hop neighbors (i.e., extended graph), which has 10.5 million nodes and 41.3 million edges. As illustrated in the Tab.~\ref{tab:inference-latency-khop}, as the neighbor hops increase, our RT-Net architecture achieves larger inference speedups.
For example, LNN on CPU devices for the 6-hop neighbor graph is 1335.98 ms, while RT Net only costs 58.44 ms --- a
22.86$\times$ improvement.




\subsection{\textbf{\textcolor{black}{Discussion on Using BRIGHT in Other Production Scenarios}}}

\textcolor{black}{We can use the Two-Stage Directed Graph in scenarios such as fraud detection for payment transaction, where we require constant update of the transaction snapshots (the frequency should be less than one day). In this use case, the current design in BRIGHT would not be optimal, as Batch Net embeddings are not updated frequently enough due to the low update frequency of data warehouse. We can mitigate this problem by the integration with graph database in data warehousing. }		

\section{Related work}
\label{sec:related}
We discuss previous work that is relevant to BRIGHT.

\textbf{(Graph Neural Network.)} 
GNN \cite{Hamilton2017InductiveRL,Kipf2017SemiSupervisedCW,Vaswani2017AttentionIA} has become increasingly popular in learning from graphs. Through message passing, aggregation and self-attention, it has a powerful capacity to grasp the graph structure, as well as the complex relations between nodes. Recent work on GNN has also focused on heterogeneous graphs \cite{hu2020heterogeneous, lv2021we}, where nodes and edges can be of various types. 

\textbf{(Modelling Dynamics with GNN.)}    Dynamic graphs can also be represented as a sequence of time events. There are several works along this line and mostly on homogeneous graphs. Temporal Graph Networks (TGN) \cite{rossi2020temporal} applied memory modules and graph-based operators. The TGN framework is computationally efficient based on event update. Asynchronous Propagation Attention Network (APAN) \cite{Wang_2021} adopted a temporal encoding similar to TGN and decoupled graph computation and inference. However, for TGN, only a few neighbors are accessible by the graph module due to memory constraints. One typical work is DySAT \cite{sankar2020dysat} which applies self-attention networks to learn low-dimensional embeddings of nodes in a dynamic homogeneous graph. One notable difference in our setting is that we need to distinguish between two types of entities (historical and future), while DySAT assumes that all entities can be added or removed in the graph at any time point. 

\textbf{(Dynamic GNN on Fraud Detection.)}  
As discussed on Sec.~\ref{sec:priliminary}, we have done previous work on flagging frauds in a dynamic setting (DHGReg \cite{rao2020suspicious} and DyHGN \cite{rao2022attention}). However, these approaches are not efficient in online deployment due to three issues that we analyzed in Sec.~\ref{sec:priliminary}. In BRIGHT, we design modules to address these issues so that GNN is deployable in real-time.

\textbf{(Two-Tower Models.)} We have surveyed the literature on two-tower / stage models that attempt to incorporate batch and streaming processing in a Lambda architecture.
In \cite{covington2016deep, 48840googletwotower}, a two-tower
model is proposed to pair users and items in the video recommendation system. One tower is for user embedding generation, and the other tower is adopted for item embedding generation. The Dual Augmented Two-tower Model (DAT) \cite{yu2021dual} customizes an augmented vector for each query and item to tackle imbalanced category data. Attention layers are utilized in \cite{wang2021exploring} to improve the recommendation of warm- and cold-start items in online video services. Cross-Batch Negative Sampling (CBNS) techniques \cite{wang2021cross}  are used to increase training of the two-tower model. Mixed Negative Sampling (MNS) \cite{yang2020mixed} uses a mix of batch and uniform sample strategies to reduce selection bias during training for the app recommendation system.
These applications focus on deep models on recommendation systems, with user embeddings and item embeddings precalculated. We setup a similar framework in BRIGHT, the neighbor embeddings are learned by a Batch Net and stored in a key-value store, which reduces the cost in online GNN deployment.

\textbf{(Model Inference Acceleration on GNN.)} Recently, people started looking at the efficiency of online GNN inference. \cite{zhou2021accelerating} applies channel pruning to speed up real-time GNN inference, by reducing Multiplication-and-ACcumulation (MAC) operations. They have also explored precomputation and storing of node representations to boost performance. GLNN \cite{zhang2021graph} applied knowledge distillation (KD) to teach multi-layer perceptrons (MLPs) from trained GNN models. The performance of GLNN is competitive. However, the graph neighbor information is not utilized in inference. Our work focuses on relation modeling, so our model is still a vanilla GNN model. In BRIGHT, during real-time inference, neighbor information is still accessed from our precalculated entity embeddings. Our BRIGHT framework is among the first to scale dynamic GNN into an industrial dataset (>60 million nodes and >160 million edges as we show in Tab.~\ref{tab:inference-latency} for static graph) and has provided an industrial benchmark in system deployment.

\section{Conclusion}

In this paper, we explore how to apply GNN for fraud detection. 
We present a novel BRIGHT framework for efficient end-to-end learning and real-time inference without the leakage of future information.
We designed a graph transformation approach to avoid future information leakage.
Furthermore, a Lambda neural network architecture was proposed to support both batch and real-time processing.
Our empirical study showed that BRIGHT outperformed baselines by 2\% w.r.t.~average precision and significantly reduced inference latency by 7.8$\times$ on average.


\begin{acks}
CZ and the DS3Lab gratefully acknowledge the support from the Swiss State Secretariat for Education, Research and Innovation (SERI) under contract number MB22.00036 (for European Research Council (ERC) Starting Grant TRIDENT 101042665), the Swiss National Science Foundation (Project Number 200021\_184628, and 197485), Innosuisse/SNF BRIDGE Discovery (Project Number 40B2-0\_187132), European Union Horizon 2020 Research and Innovation Programme (DAPHNE, 957407), Botnar Research Centre for Child Health, Swiss Data Science Center, Alibaba, Cisco, eBay, Google Focused Research Awards, Kuaishou Inc., Oracle Labs, Zurich Insurance, and the Department of Computer Science at ETH Zurich.
\end{acks}

\bibliographystyle{ACM-Reference-Format}
\balance
\bibliography{sample-base}


\end{document}